\documentclass{article}

\usepackage[preprint]{neurips_2026}
\newcommand{\WorkshopName}{NeurIPS 2026 Workshop}
\workshoptitle{\WorkshopName}

\usepackage[utf8]{inputenc}
\usepackage[T1]{fontenc}
\usepackage{amsmath,amssymb,amsfonts,mathtools}
\usepackage{booktabs}
\usepackage{graphicx}
\usepackage{float}
\usepackage{microtype}
\usepackage{needspace}
\usepackage{xcolor}
\usepackage{url}
\usepackage{hyperref}

\hypersetup{
  colorlinks=true,
  citecolor=blue,
  linkcolor=blue,
  urlcolor=blue,
  pdftitle={Adaptive Decoding Methods for Token-Efficient Reinforcement Learning},
  pdfauthor={Anonymous Authors}
}

\title{Adaptive Decoding Methods for Token-Efficient Reinforcement Learning}

\author{%
  Zhaonan Li$^*$\footnotemark[2]\thanks{Work done during the internship at Capital One.}\\
  School of Computing and Augmented Intelligence \\
  Arizona State University \\
  \texttt{zhaonan2@asu.edu} \\
  \And
  Zelei Cheng\footnotemark[2] \\
  AI Foundations \\
  Capital One \\
  \texttt{zelei.cheng@capitalone.com} \\
  \And
  Amritansh Mishra\thanks{Equal Contribution.} \\
  AI Foundations \\
  Capital One \\
  \texttt{amritansh.mishra@capitalone.com} \\
  \And
  Chia-Hsuan Lee\\
  AI Foundations \\
  Capital One \\
  \texttt{chiahsuan.lee@capitalone.com} \\
  \And
  Mann Patel\\
  AI Foundations \\
  Capital One \\
  \texttt{mann.patel2@capitalone.com} \\
  \And
  Sambit Sahu\\
  AI Foundations \\
  Capital One \\
  \texttt{sambit.sahu@capitalone.com} \\
  \And
  Ben Zhou\\
  School of Computing and Augmented Intelligence \\
  Arizona State University \\
  \texttt{benzhou@asu.edu} \\
  \And
  William Campbell\\
  AI Foundations \\
  Capital One \\
  \texttt{william.campbell@capitalone.com} \\
}

\begin{document}
\raggedbottom

\maketitle

\begin{abstract}

Rollout decoding affects both which trajectories are used for
reinforcement learning with verifiable rewards (RLVR) and how many
completion tokens are spent collecting them. We study whether
candidate-token support should remain fixed or widen during training
in the context of rollout sampling efficiency. Our main schedule moves
from top-$K=10$ to $K=20$ and then to full support as the attempted-token
budget increases. In a 4B code-RLVR setting, the schedule has a higher
interpolated 150M-token validation value than its matched fixed-$K=10$
continuation in all three runs, with a mean gain of
$1.80{\pm}0.42$ points. The schedule also performs comparably to a separately trained
fixed-$K=20$ baseline, the strongest fixed-support setting in our sweep. Follow-up experiments suggest that late widening to
full support is the most promising variant, although the evidence remains noisy,
and mixed-reward-group diagnostics do not clearly explain the gain.
An exploratory single-run test replay yields identical point estimates
for the schedule and fixed $K=10$. Overall, our results show that
rollout support is an important control for sampling efficiency in RLVR
and motivate further study of adaptive support schedules.
\end{abstract}

\section{Introduction}
\label{sec:introduction}

RLVR trains reasoning models from sampled solutions whose outcomes can be checked automatically. In group-relative training, a prompt whose rollouts are all correct or all incorrect supplies no reward contrast, but every attempted response still consumes generation tokens. The rollout policy therefore affects both the optimization signal and the cost of obtaining it.

Candidate-token support is a direct but understudied rollout control. Full-support sampling admits the entire vocabulary, including unlikely continuations; top-$K$ sampling concentrates generation on the model's $K$ most likely next tokens. A narrow support may avoid unproductive tail samples early, but it may also suppress alternatives that become useful as the policy changes. We call training-time rules that change this support \emph{adaptive rollout decoding}; the evaluation decoder remains fixed. This creates a concrete question: should rollout support remain fixed, or widen with training progress? Intuitively, the useful support may change as the policy improves. With a limited rollout-token budget, a narrow support concentrates sampling on high-probability continuations and avoids spending tokens on unlikely tail choices. However, the same restriction can exclude lower-ranked alternatives that may contain useful solution paths. This suggests a simple hypothesis: narrower support may be useful for concentrating sampling early in training, while broader support may become more useful later as the policy improves.

We test this hypothesis under an attempted-completion-token budget. Here, \emph{token-efficient} refers to validation performance as a function of generated completion tokens. At a common 36.84M-token development budget, an observed $K=10$ value from one development run is 17.73\%, while independently trained $K=20$ and full-support trajectories interpolate to 16.18\% and 16.90\%. In the separate 150M-token primary cohort, fixed $K=20$ achieves a higher estimated mean than fixed $K=10$. These cross-cohort observations suggest that the useful rollout support may depend on training progress and motivate a simple widening schedule: $K=10$ initially, $K=20$ after a nominal 50M attempted completion tokens, and full support after a nominal 100M tokens. Temperature, nucleus cutoff, group size, optimizer, verifier, and evaluation decoder are held fixed.

The common-budget estimates favor widening over every matched fixed-$K=10$ continuation, while fixed $K=20$ remains the strongest static control. A follow-up decomposition makes late direct widening the most promising next design and shows that the exact three-stage effect varies by cohort. Mixed-reward-group diagnostics and an exploratory test replay help clarify the scope and limitations of the result. Accordingly, our contributions are (i) budget-indexed and entropy-gated adaptive decoding rules for rollout support, with the latter treated as exploratory; (ii) transparent shared-parent and static-control estimates under a common token accounting rule; and (iii) a decomposition that separates the stable empirical pattern from the next hypotheses.

\section{Related Work}
\label{sec:related}

\paragraph{Verifiable rewards and group-relative optimization.}
Cobbe et al. showed that learned verifiers can rank sampled solutions and substantially improve performance on GSM8K~\citep{cobbe2021verifiers}. GRPO estimates relative advantages from groups of responses without a separate critic~\citep{shao2024deepseekmath}. DAPO develops group-relative RLVR at scale through asymmetric clipping, token-level losses, and dynamic filtering of all-correct or all-incorrect groups~\citep{yu2025dapo}. Our optimizer follows this group-relative setting; our focus is the behavior policy that generates each group.

\paragraph{Efficient and informative rollout collection.}
Recent work improves RLVR sampling by deciding where additional responses are most useful. Gradient-variance minimization assigns prompt-specific rollout counts under a compute budget~\citep{yao2025optimizing}, while AR3PO adapts response allocation and reuses collected responses~\citep{zhang2025ar3po}. DAPO's dynamic sampler similarly avoids groups with no reward contrast~\citep{yu2025dapo}. These approaches change prompt or group allocation. We instead hold the group size at eight and control which token candidates may enter every rollout, measuring the resulting learning curve against attempted completion tokens.

\paragraph{Truncation and adaptive decoding.}
Top-$K$ and nucleus sampling truncate the low-probability tail rather than merely rescale it~\citep{holtzman2020degeneration}. Truncation can also be viewed as estimating the plausible token support, with entropy-dependent thresholds adapting that support to context~\citep{hewitt2022truncation}. Mirostat provides another feedback-controlled decoder by adjusting top-$K$ from observed token surprise~\citep{basu2021mirostat}. These methods target inference-time generation. We use support control as a training-time intervention: it changes the trajectories, rewards, and token cost seen by RLVR, while evaluation decoding remains fixed.

\paragraph{Exploration schedules and entropy.}
Multi-temperature RLVR varies temperature at token or rollout levels~\citep{zhuang2025exploring}, while difficulty-aware exploration couples adaptive rollout budgets with temperature control~\citep{liao2025enhancing}. High-entropy tokens have also been identified as sparse reasoning forks that carry disproportionate policy-gradient value~\citep{wang2025beyond}. Our budget schedule is related to curricula that change a learning condition with progress~\citep{bengio2009curriculum}, but uses actual attempted-token consumption as its clock. Temperature remains fixed: the schedule changes token admissibility, and the entropy gate decides where widening is applied. The next section connects these controls to group-relative learning in one training loop.

\section{Method}
\label{sec:method}

\subsection{Method overview}
\label{sec:method_overview}

Our method adapts the candidate-token support used to collect RLVR rollouts. At collection batch $j$, the accumulated attempted-completion-token budget $B_j$ selects the nominal support: $K=10$ before 50M tokens, $K=20$ from 50M to 100M, and full support thereafter. The optional entropy-gated variant applies widening selectively at each token, retaining $K=10$ when entropy of the renormalized top-20 distribution is below $0.25$ nats and otherwise using the nominal wider support. We sample eight completions per prompt from the resulting behavior policy and assign binary code-execution rewards.

Within each prompt group, the verified rewards produce group-relative advantages. The actor is then updated with the correction-weighted, dual-clipped PPO objective and reference-KL penalty. Every generated completion token, including tokens from incorrect and length-truncated responses, increments $B_j$ and advances the support schedule. Thus adaptive collection, verification, optimization, and resource accounting share one token clock. The following subsections define each component and explain why rollout diversity determines whether a group supplies reward-driven gradient signal.

\subsection{Group-relative learning and the token clock}
\label{sec:problem_setup}

Let $x\sim\mathcal D$ be a prompt, $y_i=(a_{i,1},\ldots,a_{i,L_i})$ one of $G=8$ sampled completions, and $r_i\in\{0,1\}$ its code-execution reward. The group-relative advantage is
\begin{equation}
  \widehat A_i=\frac{r_i-\bar r}{s_r+\epsilon_A},
  \qquad
  \bar r=\frac{1}{G}\sum_{j=1}^{G}r_j,
  \label{eq:group_advantage}
\end{equation}
where $s_r$ is the within-group sample standard deviation and $\epsilon_A=10^{-6}$. Thus an all-correct or all-incorrect group has zero reward-driven relative advantage.

For the binary outcome rate $q_x^e$ induced by rollout rule $e$, conditionally independent samples give
\begin{equation}
  P_{\mathrm{mixed}}(x;e)=1-(q_x^e)^G-(1-q_x^e)^G.
  \label{eq:mixed_probability}
\end{equation}
This identity motivates recording mixed-reward groups, but it makes no directional prediction: a sampler can create contrast by producing additional failures. We therefore treat the mixed-group rate as a diagnostic and validation learning as the outcome.

Because support can also affect response length, we index training by attempted completion tokens rather than optimizer steps. If $\mathcal R_{\leq j}$ contains every response attempted through collection batch $j$, then
\begin{equation}
  B_j=\sum_{y\in\mathcal R_{\leq j}} |y|.
  \label{eq:token_budget}
\end{equation}
The count covers generated completion tokens, including incorrect and length-truncated outputs, and serves as our rollout-efficiency measure.

\subsection{Reward Contrast in Group-Relative Learning}
\label{sec:gradient_perspective}

Group-relative optimization requires within-group reward contrast for
reward-driven learning. Under output-only rewards,
Mishra et al.~\citep{mishra2026policy} show that the variance of an
individual rollout's centered reward is proportional to the reward
variance under the rollout policy. Consequently, when all rollouts for
a prompt receive the same reward, their group-relative advantages vanish
and the prompt contributes no reward-driven policy-gradient signal.
We summarize the connection to our standardized GRPO objective in
Appendix~\ref{app:gradient_signal}.

Candidate support changes the rollout distribution and can therefore
change the probability of obtaining mixed-reward groups. At the same
time, wider support can change response length and the
support-dependent correction weights in our objective. We therefore
view support adaptation as a hypothesis about the trade-off between
obtaining reward contrast and the token cost of rollout collection,
rather than assuming that wider support is always beneficial.
Accordingly, we use mixed-reward groups as a mechanism diagnostic and
validation performance per attempted token as the primary outcome.

\subsection{Support-controlled rollout decoding and optimization}
\label{sec:sampler}

Let $z_{t,a}$ be the old policy's logit for token $a$ at state $s_t$, and let $\mathcal C_K(z_t)$ contain the $K$ largest-logit tokens, with $\mathcal C_{\mathrm{full}}(z_t)=\mathcal V$. The behavior distribution is
\begin{equation}
  \mu_{\mathrm{old}}^K(a\mid s_t)=
  \frac{\mathbf 1\{a\in\mathcal C_K(z_t)\}\exp(z_{t,a})}
  {\sum_{b\in\mathcal C_K(z_t)}\exp(z_{t,b})}.
  \label{eq:support_policy}
\end{equation}
All training arms use $T=1$ and $p=1$, so changing $K$ changes admissibility without changing relative log-odds among retained tokens.

For sampled tokens, the policy ratio is $\rho_{i,t}=\pi_\theta(a_{i,t}\mid s_{i,t})/\pi_{\mathrm{old}}(a_{i,t}\mid s_{i,t})$. Let
\begin{align}
  u(\rho,A)&=\max\{-\rho A,-\operatorname{clip}(\rho,1-\epsilon,1+\epsilon)A\},\\
  d(\rho,A)&=
  \begin{cases}
    u(\rho,A), & A\geq0,\\
    \min\{-cA,u(\rho,A)\}, & A<0,
  \end{cases}
  \label{eq:dual_clip}
\end{align}
with $\epsilon=0.2$ and $c=3$. The first line is the negative per-sample clipped PPO surrogate---the minimization-loss form of the PPO clipped objective~\citep{schulman2017ppo}; the second is dual clipping~\citep{ye2020mastering}. The rollout engine records the processed log probability $\ell^{\mathrm{roll}}_{i,t}$, while the actor recomputes the old full-softmax log probability $\ell^{\mathrm{actor}}_{i,t}$. We form
\begin{equation}
  w_{i,t}=\min\!\left\{2,\exp(\ell^{\mathrm{actor}}_{i,t}-\ell^{\mathrm{roll}}_{i,t})\right\},
  \qquad
  \mathcal L_{\mathrm{pg}}=
  \frac{\sum_{i,t}M_{i,t}w_{i,t}d(\rho_{i,t},\widehat A_i)}
       {\sum_{i,t}M_{i,t}},
  \label{eq:rollout_correction}
\end{equation}
where $M$ masks padding and $w$ is detached. This matches the implemented order: PPO and dual clipping precede the unnormalized rollout weight. Under an ideal synchronized implementation, $\exp(\ell^{\mathrm{actor}}-\ell^{\mathrm{roll}})$ for a retained top-$K$ action equals the top-$K$ mass and cannot exceed one. Separately computed actor and rollout-engine likelihoods can yield larger corrections, which are capped at two. For numerical stability, both log-ratios are clamped to $[-20,20]$ before exponentiation; the low-variance KL token term is additionally clamped to $[-10,10]$. The loss includes that reference-KL penalty with coefficient $10^{-3}$. Top-$K$ truncates the collection distribution rather than the actor's parameterized softmax, thereby affecting both collected actions and their correction weights. The correction applies to sampled actions within the behavior support, and the resulting support-dependent weights can also change effective gradient scale.

\subsection{Adaptive decoding by token budget and entropy}
\label{sec:scheduler}

We evaluate the monotone schedule
\begin{equation}
  K(B)=
  \begin{cases}
    10, & 0\leq B<50\mathrm{M},\\
    20, & 50\mathrm{M}\leq B<100\mathrm{M},\\
    \mathrm{full}, & B\geq100\mathrm{M}.
  \end{cases}
  \label{eq:support_schedule}
\end{equation}
The active value is fixed for an entire collection batch; the first batch beginning at or after a threshold uses the new support. Boundaries are therefore nominal and can be exceeded by one batch. This rule supplies budget-indexed adaptation, while the entropy gate below supplies state-dependent adaptation.

The budget schedule widens support globally as training progresses, but uncertainty is not uniform across token positions. At a low-entropy position, the next-token distribution is strongly peaked, so widening the candidate set may mostly expose low-probability alternatives. At a high-entropy position, several continuations are competitive, making broader support more likely to expose meaningful alternative paths. We therefore ask whether widening can be applied selectively at locally uncertain positions. As an exploratory extension, we also gate widening at the token level. We compute the entropy $H_{20}(s_t)$ of the normalized top-20 distribution and retain $K=10$ when $H_{20}(s_t)<0.25$ nats; otherwise we use the arm's nominal wider support. This single-run experiment tests the feasibility of state-dependent support adaptation.

We compare these budget-indexed and entropy-gated decoding rules with fixed-support controls under the common training and evaluation protocol described next.

\section{Experimental Design}
\label{sec:eval}

\paragraph{Model, data, and verifier.}
We train \texttt{Llama-3.1-Nemotron-Nano-4B-v1.1} \citep{bercovich2025llama} on a difficulty-filtered subset derived from rStar-Coder~\citep{liu2025rstar}. We first sample the same base model five times per candidate problem at $T=0.6$ and $p=0.95$ (maximum 16,384 tokens), and retain problems solved in at most two of the five attempts. We focus on these low-solvability problems to stress-test rollout sampling efficiency. On easy problems, useful solutions are already sampled frequently, leaving less room for the rollout distribution to matter. On harder problems, successful trajectories are relatively rare, so how the fixed rollout-token budget is allocated can have a larger effect on whether training observes informative successes and mixed-reward groups. A fixed stratified split yields 1,323 training, 300 validation, and 300 test problems with disjoint identifiers; validation and test each contain 100 problems from the $0/5$, $1/5$, and $2/5$ strata. A single user message requests a fenced Python program, and we execute the last fenced code block using SandboxFusion~\citep{bytedanceseedfoundationcodeteam2025fullstackbenchevaluatingllms}. Reward is one only if the program passes all supplied test cases within a 10-second per-test timeout; missing code, execution errors, and timeouts receive zero.

\paragraph{Training and evaluation.}
Each update samples 64 prompts and eight responses per prompt, with maximum prompt and response lengths of 2,048 and 24,576 tokens. We use a \texttt{verl} \citep{sheng2024hybridflow} fork with vLLM collection and FSDP training, bfloat16, one policy epoch, AdamW with constant learning rate $5\times10^{-6}$, $0.01$ weight decay, $(0.9,0.999)$ betas, no warmup, gradient-norm clipping at 1, a 64-sequence PPO minibatch, and no entropy bonus. Runs use one node with eight H100 GPUs. Unless noted otherwise, $\pm$ denotes the sample standard deviation across three independent training runs.

Validation is configured every 20 optimizer updates. Because token use per update is large and variable, the resulting observations are irregular and often far apart. At each validation point, we sample five
responses per problem and average their binary outcomes.
We use the same evaluation decoder for every training arm so that differences in validation performance reflect the learned policies rather than different test-time sampling rules. The temperature and nucleus settings ($T=0.6$ and $p=0.95$) follow the model's recommended sampling configuration, while $K=20$ is held fixed across all evaluations.
The average binary outcome is the same estimator later called pass@1. With $B^\star=150$M, we report piecewise-linear estimates
\begin{align}
  \widehat{\mathrm{Acc}}_m@B^\star &= \widehat{\mathrm{ValAcc}}_m(B^\star),
  &
  \widehat{\mathrm{AUC}}_{m,B^\star} &= \frac{1}{B^\star}
    \int_0^{B^\star}\widehat{\mathrm{ValAcc}}_m(B)\,\mathrm dB,
  \label{eq:efficiency_metrics}
\end{align}
The endpoint measures performance at the target token budget, whereas
AUC captures how quickly a method acquires validation performance over
the full training horizon. The hat denotes linear interpolation between
actual evaluations; AUC uses trapezoidal integration of those segments. For a resumed comparison beginning at shared-parent budget $B_0$, post-parent AUC is $(B^\star-B_0)^{-1}\int_{B_0}^{B^\star}\widehat{\mathrm{ValAcc}}_m(B)\,\mathrm dB$. Figure~\ref{fig:token_budget_curves} exposes the raw knots, Table~\ref{tab:main_results} reports bracket widths, and both quantities are labeled as common-budget interpolation estimates. Because validation checkpoints are sparse and irregular, these quantities are descriptive common-budget estimates rather than direct observations at $B^\star$. Reported standard deviations quantify run-to-run variation conditional on the piecewise linear interpolation and do not account for uncertainty due to unobserved learning-curve shape between checkpoints.

\paragraph{Cohorts and controls.}
The primary cohort compares full support, fixed $K=10$, fixed $K=20$, and Equation~\eqref{eq:support_schedule}. The scheduled and fixed-$K=10$ arms share the model, optimizer, and data state at their within-run branch point; full support and fixed $K=20$ are independently trained static controls with separate parents and aligned replicate indices. Owing to whole-batch switching, the primary scheduled runs actually enter $K=20$ at 51.16--52.49M tokens and full support at 101.12--104.85M. A follow-up decomposition branches four continuations from each early $K=10$ parent: remain at $K=10$, change only to $K=20$, change directly to full support at 100M, or use both transitions. Two decomposition runs are new continuations, while the third reuses the corresponding primary fixed and scheduled runs. The entropy-gate comparison uses a single shared parent and is reported separately.

For the final 100--150M interval, we additionally aggregate the fraction of prompt groups whose eight rollouts contain both reward values, the mean response length, and mixed groups per million attempted tokens. Groups and tokens from batches crossing the interval boundaries are prorated. These statistics connect Equation~\eqref{eq:mixed_probability} to the observed rollouts as a mechanism diagnostic.

\section{Results}
\label{sec:results}

\subsection{Token-budgeted adaptive decoding versus fixed support}
We first ask whether widening rollout support improves validation performance at a common attempted-token budget. Our primary comparison is between the widening schedule and its matched fixed-$K=10$ continuation, which share a parent within each run. Fixed $K=20$ and full support provide independently trained static baselines. Table~\ref{tab:main_results} reports both endpoint performance and performance integrated over the 150M-token training horizon.

\begin{table}[H]
    \caption{
    Piecewise-linear common-budget validation estimates at $B^\star = 150$M attempted completion tokens (percentages; three runs per method). Reported $\pm$ values are sample standard deviations across runs conditional on the interpolation and do not capture uncertainty from sparse evaluation checkpoints. The bold row reports the matched schedule--fixed-$K=10$ difference. ``Max bracket'' is the widest run level interval between observed validation checkpoints surrounding 150M.
    }
  \centering
  \small
  \setlength{\tabcolsep}{3.5pt}
  \begin{tabular*}{\linewidth}{@{\extracolsep{\fill}}llccc@{}}
    \toprule
    \textbf{Method} & \textbf{Training support} & $\widehat{\mathbf{Acc}}\mathbf{@150M}$ & $\widehat{\mathbf{AUC}}_{\mathbf{150M}}$ & \textbf{Max bracket (M)} \\
    \midrule
    Full support & full & $18.34{\pm}2.51$ & $16.26{\pm}1.80$ & $105.95$ \\
    Fixed support & $10$ & $21.49{\pm}2.58$ & $18.56{\pm}1.50$ & $117.11$ \\
    Fixed support & $20$ & $23.35{\pm}1.28$ & $19.16{\pm}0.89$ & $42.92$ \\
    Scheduled & $10\!\rightarrow\!20\!\rightarrow\!\mathrm{full}$ & $23.29{\pm}2.51$ & $19.16{\pm}1.58$ & $102.48$ \\
    \addlinespace
    \textbf{Matched gain} & schedule $-$ fixed $K=10$ & $\mathbf{+1.80\pm0.42}$ & $\mathbf{+0.60\pm0.08}$ & --- \\
    \bottomrule
  \end{tabular*}
  
  \label{tab:main_results}
\end{table}

\begin{figure}[H]
  \centering
  \includegraphics[width=0.92\linewidth]{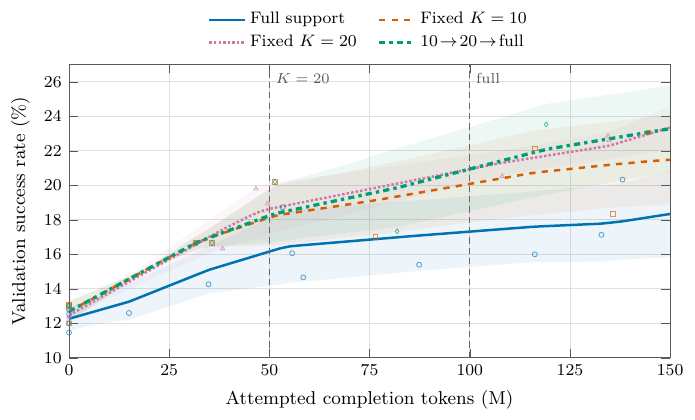}
  \caption{Validation versus attempted completion tokens for the primary support strategies. Lines and bands show mean $\pm1$ sample standard deviation after piecewise-linear interpolation; symbols mark actual evaluations. Vertical lines are nominal transitions; the $y$-axis is truncated. Fixed $K=20$ is independently trained.}
  \label{fig:token_budget_curves}
\end{figure}

\Needspace{6\baselineskip}
\paragraph{Scheduling versus the matched narrow continuation.}
Under the stated interpolation, the schedule is above its matched fixed-$K=10$ continuation in \textbf{every run}. Estimated endpoint differences are 2.12, 1.32, and 1.95 points across the three matched runs, giving $\mathbf{+1.80\pm0.42}$ points; estimated paired AUC differs by $\mathbf{+0.60\pm0.08}$ points. The shared parents make this the cleanest comparison in the study. Per-run values and checkpoint brackets are reported in Appendix~\ref{app:estimation}.

\paragraph{A tuned static support remains competitive.}
Fixed $K=20$ reaches an estimated $23.35{\pm}1.28$\% endpoint success, while the schedule reaches $23.29{\pm}2.51$\%. Corresponding-run schedule--$K=20$ differences are $-0.07{\pm}3.64$ endpoint points and $0.01{\pm}2.17$ AUC points. The study does not separate these independently trained arms, and fixed $K=20$ remains competitive. Thus the primary result supports widening over remaining at $K=10$, but does not show an advantage over a well-chosen fixed support. Figure~\ref{fig:token_budget_curves} displays both the interpolation and the observation density underlying it.

\Needspace{7\baselineskip}
\paragraph{Exploratory test replay.}
We additionally evaluate a single replay using the fixed-$K=10$ and
scheduled continuations from Table~\ref{tab:schedule_analysis}, together
with a full-support checkpoint. At the 150M-token budget, estimated
validation success is 24.05\%, 24.40\%, and 25.28\% for full support,
fixed $K=10$, and the scheduled continuation, respectively. On the
disjoint 300-problem test split, however, pass@1 is 23.07\%, 23.87\%,
and 23.87\%, respectively, so the scheduled and fixed-$K=10$
continuations have identical test point estimates. Their test
difference is 0.00 percentage points, with a paired-by-question
cluster-bootstrap interval of $[-2.80, 2.87]$ points, conditional on
the five sampled generations per problem. We treat this result as
exploratory because it is based on a single replay and because the
test split was also used in an earlier 500M analysis.

\subsection{Which transition matters?}
\label{sec:schedule_analysis}
The primary experiment shows that the full widening schedule improves over its matched $K=10$ continuation, but it does not tell us when widening helps. We therefore run a matched-parent decomposition that isolates the two support transitions: $10\!\rightarrow\!20$ and $10\!\rightarrow\!\mathrm{full}$. Separately, we use a single-run entropy-gated experiment to ask whether widening needs to occur at every token position. Two of the three decomposition runs are new stochastic continuations from their earlier parents, while the third reuses the corresponding primary run; accordingly, baseline values need not match Table~\ref{tab:main_results}.


\begin{table}[H]
    \caption{Piecewise-linear follow-up estimates at 150M attempted completion tokens. Boldface marks the highest accuracy and AUC within each comparison block. Accuracy and AUC are percentages; actor entropy is the linearly estimated logged token-mean in nats, while $H_{20}$ is the gate statistic. The decomposition reports mean $\pm$ sample standard deviation. Gate rows are single-run exploratory continuations.}
  \centering
  \small
  \setlength{\tabcolsep}{3.5pt}
  \begin{tabular*}{\linewidth}{@{\extracolsep{\fill}}lcccc@{}}
    \toprule
    \textbf{Continuation or gate} & $\mathbf{n}$ & $\widehat{\mathbf{Acc}}\mathbf{@B^\star}$ & $\widehat{\textbf{Post-parent AUC}}$ & \textbf{Actor entropy} \\
    \midrule
    \multicolumn{5}{l}{\emph{Matched-parent decomposition}} \\
    $K=10$ throughout & 3 & $22.97{\pm}3.94$ & $20.38{\pm}2.68$ & $0.348{\pm}0.013$ \\
    $10\!\rightarrow\!20$ & 3 & $21.41{\pm}2.78$ & $19.70{\pm}1.99$ & $0.359{\pm}0.043$ \\
    $10\!\rightarrow\!\mathrm{full}$ at 100M & 3 & $\mathbf{24.23\pm1.42}$ & $\mathbf{20.96\pm0.97}$ & $0.437{\pm}0.074$ \\
    $10\!\rightarrow\!20\!\rightarrow\!\mathrm{full}$ & 3 & $23.35{\pm}2.54$ & $20.44{\pm}2.03$ & $0.447{\pm}0.067$ \\
    \midrule
    \multicolumn{5}{l}{\emph{Exploratory entropy gate}} \\
    Ungated schedule & 1 & $24.30$ & $21.55$ & $0.440$ \\
    Gated, nominal $K=20$ & 1 & $24.71$ & $20.66$ & $0.392$ \\
    Gated, nominal full & 1 & $23.27$ & $21.60$ & $0.439$ \\
    Gated schedule & 1 & $\mathbf{25.62}$ & $\mathbf{21.74}$ & $0.428$ \\
    \bottomrule
  \end{tabular*}
  \label{tab:schedule_analysis}
\end{table}

Direct late expansion has the highest estimated mean in the decomposition, but the uncertainty is material: relative to its matched fixed continuation, its endpoint change is $1.26{\pm}3.06$ points (positive in two of three runs) and its post-parent AUC change is $0.58{\pm}1.88$ points (positive in one of three). Its fixed-$K=20$ comparator comes from a separate lineage. We therefore treat late widening as a promising follow-up hypothesis rather than an established improvement. The exact three-stage schedule is more consistent against the middle-only continuation ($\mathbf{+1.94\pm0.60}$ endpoint points, \textbf{positive in all three runs}), but changes by only $0.38{\pm}1.87$ points versus fixed $K=10$ in this cohort. Sparse validation observations also prevent localizing these differences sharply to a transition boundary.


In the single-run comparison, the gated schedule has a higher estimated endpoint ($25.62\%$) than the ungated schedule ($24.30\%$), while its estimated average actor entropy is lower. This single-run result motivates replicated evaluation with direct gate-activation logging.


\subsection{Do wider supports create more contrast-bearing groups?}
\label{sec:group_diagnostics}
We next ask whether wider support improves learning by producing more contrast-bearing groups for the same rollout-token budget. If this mechanism explains the validation gains, we would expect widening to increase the number of mixed-reward groups obtained per attempted token. We measure this over the final 100--150M-token interval.

\begin{table}[H]
    \caption{Rollout-group diagnostics over the final 100--150M-token interval
(three runs per method). ``Mixed'' means that a prompt's eight
rollouts contain both correct and incorrect attempts. Wider-support arms have slightly
more mixed groups but also longer completions, yielding no clear
increase in mixed groups per attempted token.}
  \centering
  \small
  \setlength{\tabcolsep}{5pt}
  \begin{tabular*}{\linewidth}{@{\extracolsep{\fill}}lccc@{}}
    \toprule
    \textbf{Final-interval support} & \textbf{Mixed groups (\%)} & \textbf{Mean length} & \textbf{Mixed groups/M tokens} \\
    \midrule
    Fixed $K=10$ & $52.33{\pm}5.08$ & $8{,}152{\pm}1{,}527$ & $8.13{\pm}0.91$ \\
    Direct late full & $54.44{\pm}3.64$ & $9{,}121{\pm}855$ & $7.48{\pm}0.36$ \\
    Three-stage schedule & $53.90{\pm}1.98$ & $8{,}421{\pm}1{,}463$ & $8.13{\pm}1.13$ \\
    \bottomrule
  \end{tabular*}
  \label{tab:group_diagnostics}
\end{table}

Relative to fixed $K=10$, the mixed-group fraction changes by $+2.11{\pm}3.67$ points for direct late expansion and $+1.56{\pm}3.19$ points for the three-stage schedule. The wider supports also produce longer completions: direct late expansion yields $-0.66{\pm}0.92$ mixed groups per million tokens, while the schedule's mean change has magnitude below $0.01$ (SD $0.37$). Thus support changes the rollout distribution, but these results do not show that widening increases the number of contrast-bearing groups obtained per attempted token.


\section{Discussion}
\label{sec:discussion}
Our results provide initial evidence that rollout support can influence
token-budgeted RLVR learning dynamics. In the primary cohort, the widening
schedule achieves a higher estimated 150M validation value than its matched
fixed-$K=10$ continuation in all three runs, while performing comparably to
the independently trained fixed-$K=20$ baseline. The follow-up decomposition
highlights late widening to full support as the most promising variant, with
some variability across runs that motivates further replication.
Mixed-reward-group diagnostics do not fully account for the observed
validation differences, leaving the mechanism of support adaptation open.

Several limitations make these conclusions preliminary. We study one
4B model in one code-RLVR setting, the primary results use three runs,
and validation checkpoints are sparse. The entropy-gating and test
replay analyses use only one run. In addition, changing support affects
both the sampled trajectories and the rollout-correction weights, so we
cannot attribute the observed differences solely to exploration.
Denser evaluations, replicated late-widening experiments, and a fresh
held-out test set are the most useful next steps.




\Needspace{8\baselineskip}
\section{Conclusion}
\label{sec:conclusion}


In this 4B code-RLVR setting, rollout support affects validation
performance as a function of attempted completion tokens. In the
primary cohort, budget-indexed widening has a higher interpolated
150M-token validation value than its matched fixed-$K=10$ continuation
in all three runs, while its estimated mean is comparable to the
independently trained fixed-$K=20$ baseline. An exploratory test replay
yields identical point estimates for the scheduled and fixed-$K=10$
continuations.

The follow-up decomposition suggests that late widening to full support
is the most promising component of the schedule, although the current
evidence is not sufficient to establish a consistent improvement.
Mixed-reward-group diagnostics also do not clearly explain the observed
validation gain, leaving the mechanism of support adaptation open.
More broadly, our results suggest that candidate support should be
treated as an important rollout hyperparameter in RLVR rather than
defaulting to either narrow or full support. Adaptive strategies should
therefore be evaluated over the full training horizon, against strong
fixed-support baselines, and under the rollout-token budget that they
directly affect.

\begingroup
\bibliographystyle{plainnat}
\bibliography{neurips_2026_references}
\endgroup

\appendix

\section{Connection Between Reward Contrast and GRPO Gradient Signal}
\label{app:gradient_signal}

Our motivation builds on the gradient analysis of
Mishra et al.~\citep{mishra2026policy}. We specialize the relevant
result to our binary-reward setting and clarify its relationship to the
standardized advantages used in our implementation.

For a prompt $x$, let $r_i \in \{0,1\}$ be the terminal reward of
rollout $i$, let
\[
    \bar r = \frac{1}{G}\sum_{j=1}^{G}r_j,
\]
and define the centered reward
\[
    U_i = r_i-\bar r.
\]
If the $G$ rollouts are conditionally independent with success
probability $q_x^e$ under rollout rule $e$, then
\begin{equation}
    \operatorname{Var}(U_i\mid x,e)
    =
    \frac{G-1}{G}\sigma_{r,x,e}^{2},
    \qquad
    \sigma_{r,x,e}^{2}
    =
    q_x^e(1-q_x^e).
    \label{eq:appendix_reward_variance}
\end{equation}
This is the binary-reward specialization of the advantage-variance
result of Mishra et al.~\citep{mishra2026policy}. By Chebyshev's
inequality,
\begin{equation}
    \Pr(|U_i|>\varepsilon\mid x,e)
    \le
    \frac{(G-1)\sigma_{r,x,e}^{2}}
         {G\varepsilon^{2}}.
\end{equation}

As $q_x^e$ approaches $0$ or $1$, reward variance decreases and
mixed-reward groups become rare. In the limiting case of a uniform
group, $r_1=\cdots=r_G$, we have $U_i=0$ for every rollout. Our
standardized advantage,
\[
    \widehat A_i
    =
    \frac{r_i-\bar r}{s_r+\epsilon_A},
\]
is therefore also exactly zero, so the reward-driven PPO surrogate
receives no contribution from that group.

\section{Per-Run Evaluation Details}
\label{app:estimation}

Table~\ref{tab:per_run_results} gives the run-level quantities summarized in Table~\ref{tab:main_results}; Table~\ref{tab:endpoint_brackets} gives the raw checkpoints used to estimate the 150M endpoint.

\begin{table}[h]
    \caption{Per-run piecewise-linear estimates in percentage points at 150M attempted completion tokens.}
  \centering
  \small
  \setlength{\tabcolsep}{4pt}
  \begin{tabular*}{\linewidth}{@{\extracolsep{\fill}}lccc@{}}
    \toprule
    \textbf{Method} & \textbf{Run A Acc/AUC} & \textbf{Run B Acc/AUC} & \textbf{Run C Acc/AUC} \\
    \midrule
    Full support & 20.95 / 18.28 & 15.95 / 14.83 & 18.14 / 15.67 \\
    Fixed $K=10$ & 23.17 / 19.79 & 22.77 / 19.01 & 18.52 / 16.89 \\
    Fixed $K=20$ & 21.90 / 18.17 & 23.83 / 19.88 & 24.33 / 19.42 \\
    Scheduled & 25.29 / 20.47 & 24.10 / 19.62 & 20.47 / 17.41 \\
    \bottomrule
  \end{tabular*}
  \label{tab:per_run_results}
\end{table}

\begin{table}[h]
    \caption{Actual evaluation checkpoints immediately bracketing 150M attempted completion tokens. As a descriptive check, the mean success rates at each run's nearest raw checkpoint are 17.80, 21.18, 23.62, and 23.20\% in row order; those checkpoints occur at different token counts.}
  \centering
  \small
  \setlength{\tabcolsep}{4pt}
  \begin{tabular*}{\linewidth}{@{\extracolsep{\fill}}lccc@{}}
    \toprule
    \textbf{Method} & \textbf{Run A (M)} & \textbf{Run B (M)} & \textbf{Run C (M)} \\
    \midrule
    Full support & 138.1--244.1 & 116.2--158.4 & 132.8--225.2 \\
    Fixed $K=10$ & 144.6--261.7 & 116.2--221.9 & 135.7--208.8 \\
    Fixed $K=20$ & 108.1--151.0 & 134.5--153.9 & 134.8--154.7 \\
    Scheduled & 51.4--153.5 & 119.1--221.5 & 81.9--152.8 \\
    \bottomrule
  \end{tabular*}
  \label{tab:endpoint_brackets}
\end{table}

\section{Broader Impacts}
\label{app:impact}

Reducing attempted completion tokens can lower one component of RLVR cost, while energy use remains a system-level property. Model, training-data, and deployment risks continue to require task-specific safety evaluation. More capable reasoning models may be used in harmful applications, and verifier-based training can amplify specification errors when the verifier rewards an incomplete proxy for the intended behavior.

\end{document}